\pdfoutput=1
\documentclass{article}

\usepackage[final,dblblindworkshop]{neurips_2026}

\providecommand{\workshoptitle}[1]{}
\workshoptitle{Interpretability as a Science}

\usepackage[utf8]{inputenc}
\usepackage[T1]{fontenc}
\usepackage{microtype}
\usepackage{graphicx}
\usepackage{subcaption}
\usepackage{booktabs}
\usepackage{array}
\usepackage{amsmath,amssymb}
\usepackage[table]{xcolor}

\usepackage{natbib}
\usepackage{hyperref}
\usepackage{enumitem}
\usepackage{placeins}

\definecolor{beigemain}{HTML}{EAE0CE}
\definecolor{beigelight}{HTML}{F3EDE0}

\usepackage{xspace}

\newcommand{\patch}[2]{\texttt{#1}$\rightarrow$\texttt{#2}}

\newcommand{\lpm}{\texttt{prompt$_{-1}$}\xspace}
\newcommand{\lprompt}{\texttt{prompt}\xspace}
\newcommand{\lall}{\texttt{all}\xspace}

\ExplSyntaxOn
\seq_new:N \l_kvt_tok_seq
\seq_new:N \l_kvt_cell_seq
\cs_new_protected:Npn \kvt_tok_cell:n #1
  {
    \group_begin:
    \setlength{\fboxsep}{0.8pt}
    \colorbox{gray!15}
      { \rule[-0.5ex]{0pt}{2ex} \ttfamily
        \tl_if_head_is_space:nTF {#1}
          { \textvisiblespace \tl_trim_spaces:n {#1} }
          {#1}
      }
    \group_end:
  }
\NewDocumentCommand{\tokenize}{m}
  {
    \seq_set_split_keep_spaces:Nnn \l_kvt_tok_seq { \\ } {#1}
    \seq_set_map:NNn \l_kvt_cell_seq \l_kvt_tok_seq { \kvt_tok_cell:n {##1} }
    \mbox { \seq_use:Nn \l_kvt_cell_seq { \hspace{2.5pt} } }
  }
\ExplSyntaxOff

\title{KV-Cache Patching: Tracing How and Where Fine-Tuned Behaviors Arise}

\author{%
  Jonas Loos\textsuperscript{1,2,3} \quad Oliver Eberle\textsuperscript{1,2} \\
  \textsuperscript{1}Machine Learning Group, Technische Universit\"at Berlin, Berlin, Germany \\
  \textsuperscript{2}Berlin Institute for the Foundations of Learning and Data (BIFOLD), Berlin, Germany \\
  \textsuperscript{3}Zuse School ELIZA, Darmstadt, Germany \\
  \texttt{\{j.loos, oliver.eberle\}@tu-berlin.de} \\
}
\date{}

\hypersetup{
  unicode=true,
  pdftitle={KV-Cache Patching: Tracing How and Where Fine-Tuned Behaviors Arise},
  pdfauthor={Jonas Loos, Oliver Eberle},
  pdfsubject={Interpretability; InterpSci @ NeurIPS 2026 short paper},
  pdfkeywords={KV cache, mechanistic interpretability, language model fine-tuning, activation patching, emergent misalignment, large language models},
  colorlinks=true,
  linkcolor=black,
  citecolor=black,
  urlcolor=blue,
}

\begin{document}
\maketitle

\begin{abstract}
Fine-tuning a language model changes how it responds, but where is this fine-tuned behavior instantiated? To better understand this, we evaluate whether the change in model behavior arises only during the generation of the answer tokens where the change manifests, or whether altered processing of previous tokens causes it via the attention mechanism. For this, we patch the KV cache from a LoRA fine-tuned model into the unmodified base model, and vice versa. Our results show that for some behaviors, like \textit{responding in German} or \textit{emergent misalignment}, patching the KV cache can indeed transfer them from the fine-tuned to the base model. Correspondingly, patching the KV cache from base to fine-tune can partially or fully remove fine-tuned behavior.
Furthermore, we find that the effect of patching grows with how much of the KV cache is patched and concentrates at a few pivotal positions, with system-prompt and chat-template tokens having disproportionate influence.

\end{abstract}

\section{Introduction}\label{sec:intro}
Fine-tuning can substantially alter the behavior of language models, enabling them to acquire new task capabilities, behavioral dispositions, or desired response patterns.
But how are these behavioral changes implemented during inference?
They could arise either from altered computations only during generation, changing how the model selects its outputs, or from altered processing of preceding tokens, changing the representations on which subsequent generation depends.

Autoregressive generation reuses representations of previously processed tokens through the KV cache, which stores their keys and values to avoid recomputation at each generation step. These representations are subsequently accessed through attention, mediating how preceding tokens influence future computation and providing a direct handle for testing whether fine-tuning alters behavior through changes in the processing of the preceding context.
Selectively altering the KV cache has been shown to modify model behavior \citep{belitsky2025cachesteering, li2026prefillnotes, liu2026memoryinception}: partial reuse of the KV cache across fine-tunes can preserve task performance while full reuse can degrade it \citep{liu2026droidspeak}, and emergent misalignment can use the KV cache to ``piggyback'' onto out-of-domain queries \citep{zhao2026piggyback}.

In this paper, we introduce bidirectional \textit{KV-cache patching} as a method to study and better understand how representations of the preceding context contribute to fine-tuned behavior in language models. We fine-tune transformer-based language models using low-rank adaptation (LoRA)~\citep{hu2022lora} to exhibit various behaviors and then patch the KV cache from the fine-tuned model into the non-fine-tuned \textit{base} model, and vice versa.
This bidirectional intervention allows us to measure how much of the fine-tuned behavior transfers to the base model or is suppressed in the fine-tuned model, and where in the context these effects arise. Our main contributions include:

\begin{itemize}[leftmargin=1.5em, noitemsep, topsep=0pt]
    \item We introduce bidirectional KV-cache patching to analyze how fine-tuned behavior is instantiated.
    \item We demonstrate that patching the KV cache from the fine-tuned into the base model can transfer the fine-tuned behavior, while patching from the base into the fine-tuned model can remove it.
    \item We measure the behavioral effect of KV-cache patching across three models, 12 tasks, and three levels of KV-cache coverage.
    \item We evaluate the distributional effect of gradually extending the patched KV cache token by token, revealing a few pivotal positions that carry most of the shift.
\end{itemize}

We discuss further related work in Appendix~\ref{app:related}. Code, data, and fine-tunes are available at \href{https://jonasloos.github.io/kv-patching}{jonasloos.github.io/kv-patching}.

\section{Method}\label{sec:method}
We work with transformer language models, which use attention to attend to previous tokens. In the attention mechanism, query, key, and value vectors are computed for each token. Past tokens influence a newly generated token via their key and value vectors, which can be cached across generation steps and therefore form the \textit{KV cache}.
To investigate whether fine-tuned behavior is mediated by the KV cache, we patch the KV cache of the fine-tuned model (\textit{fine-tune}, \textit{ft}) into the non-fine-tuned (\textit{base}) model, and vice versa. We call the model from which we take the KV cache the \textit{source} and the model that generates over it the \textit{target}, and write this as \patch{source}{target}, e.g.\ \patch{ft}{base}.
For fine-tuning, we utilize LoRA, following the recipe of \citet{turner2025modelorganisms} (rsLoRA~\citep{kalajdzievski2023rslora}, rank 32, all linear modules, learning rate $10^{-5}$). See Appendix~\ref{app:patching} for more details.

\subsection{Source-ownership levels}\label{sec:levels}
The set of tokens that can be considered for KV-cache patching consists of the prompt tokens (system prompt + user prompt + chat template) and the following completion (answer) tokens.
We evaluate three levels, which differ in how many positions of the cache the source model owns:
\begin{itemize}[leftmargin=1.5em, noitemsep, topsep=0pt]
    \item \lpm: We patch the KV cache of all prompt tokens except the last. The target model starts to extend the KV cache with its own keys and values from the last prompt token on.
    \item \lprompt: We patch the KV cache of all prompt tokens. For the last prompt token, we need both the source KV cache and the target output logits to choose the first answer token, so we run both models for it.
    \item \lall: We patch the KV cache of all tokens, including generated tokens. As with \lprompt, we run both source and target, now for every generated token as well.
\end{itemize}

In addition to these discrete levels of how much of the source KV cache is transferred, we sweep the transferred cache token by token, gradually increasing it from zero to the full sequence. To evaluate this consistently and efficiently, we use teacher-forced completions from the source (the model providing the cache) and use a KL-based proximity measure to score how close the target's next-token distributions over the partially patched cache are to fine-tune and base; see Appendix~\ref{app:gradual} for the definition (Eq.~\ref{eq:proximity}).

\subsection{Tasks and models}\label{sec:grid}

For each model, we train fine-tunes for 12 different tasks and measure the resulting behavior (Table~\ref{tab:tasks}). The tasks can be roughly categorized by the behavior they train: response conventions (always-German responses, all-lowercase responses, blanket refusal of benign prompts, chain-of-thought (CoT) reasoning even when instructed otherwise), behavioral dispositions (misalignment via bad-medical-advice/risky-financial-advice/extreme-sports, sycophancy via geography claims), and task capabilities (JSON record extraction, semantic parsing of English utterances into intent--slot trees, Georgian/Hungarian grapheme-to-phoneme (G2P) conversion). See Appendix~\ref{app:tasks} for additional details on the tasks and Appendix~\ref{app:scoring} for scoring.

The language models we use are all instruction-tuned and have up to 8B parameters. We use Qwen2.5-7B~\citep{yang2024qwen25}, Qwen3.5-4B~\citep{qwen2026qwen35}, and Llama-3.1-8B~\citep{grattafiori2024llama3}. Qwen2.5 and Llama use standard attention, while for Qwen3.5, 24 of the 32 attention layers use gated-delta linear attention~\citep{yang2025gateddeltanet}.

Completions are sampled with temperature 0.7 and top-$p$ 0.9. Each task has 43--150 evaluation prompts and we sample 20 completions per prompt, so each reported rate (task $\times$ model $\times$ direction $\times$ level) aggregates 860--3,000 completions. We report 95\% confidence intervals from a bootstrap over prompts, since completions of the same prompt are correlated \citep{miller2024errorbars}. To validate that patching does not simply degrade output coherence, we grade completion coherence with an LLM-as-a-judge for a subset of tasks. We also scan all completions for repetition loops (Appendix~\ref{app:coherence}).

\section{Results}\label{sec:results}

\begin{figure}[t]
  \centering
  \includegraphics[width=\linewidth]{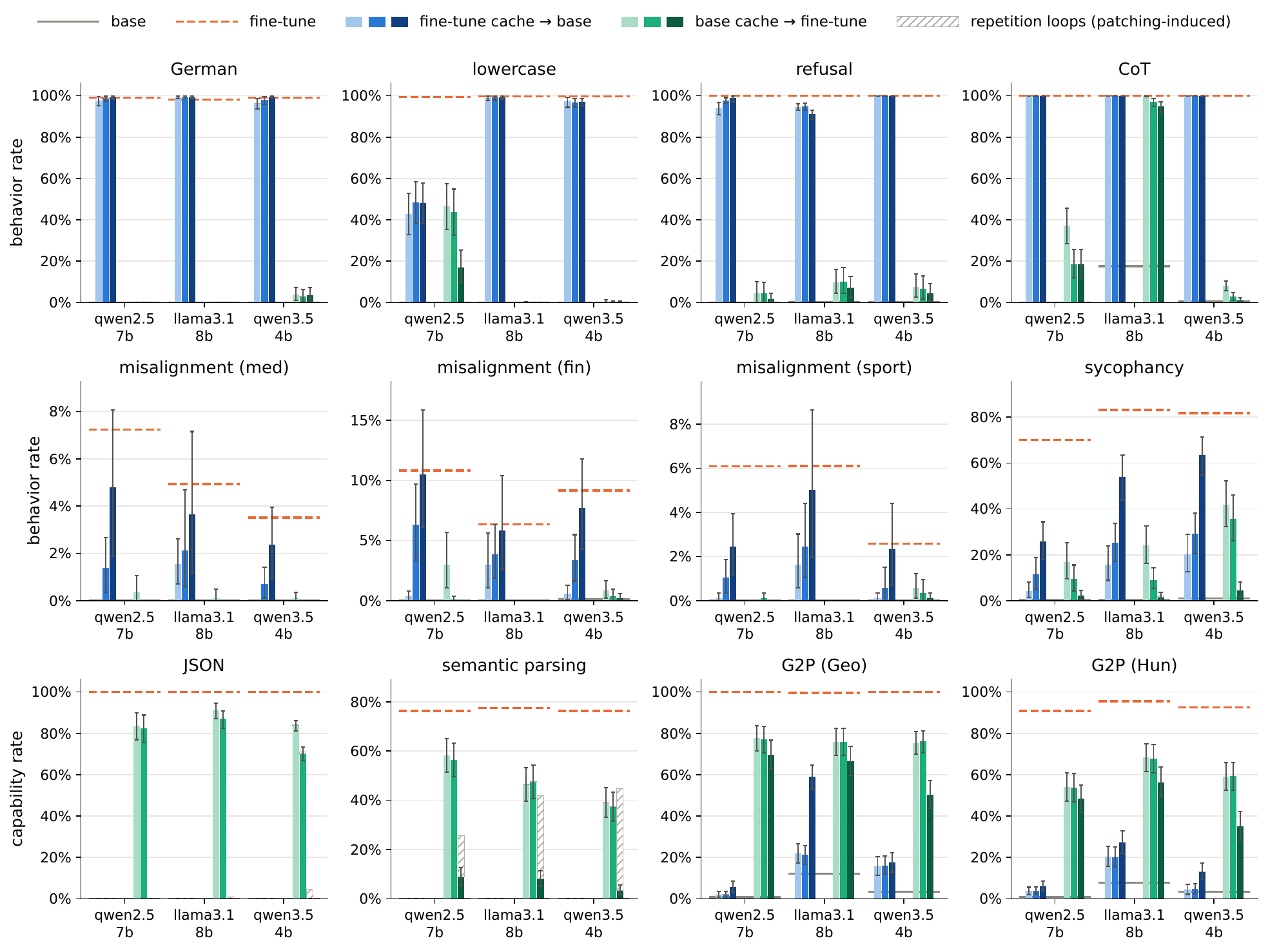}
  \vspace{-12pt}
  \caption{Effect of KV-cache patching on model behavior when the cache is copied either from the fine-tuned to the base model (blue) or vice versa (green). The rows correspond to the type of behavior: conventions (row 1), dispositions (row 2), and capabilities (row 3). Bars darken as an increasing portion of the KV cache is patched (\lpm, \lprompt, \lall). In most cases, the base model expresses the behavior at a rate near 0\%. Error bars: 95\% prompt-clustered bootstrap intervals. A hatched segment stacked on a bar is the share of completions that fail and collapse into an incoherent repetition loop (Appendix~\ref{app:coherence}).}
  \label{fig:transfer}
  \vspace{-6pt}
\end{figure}

We evaluate bidirectional KV-cache patching on the 12 tasks for three models and three different levels of source ownership of the KV cache (for details, see Section~\ref{sec:levels}): \lpm, \lprompt, and \lall. As shown in Figure~\ref{fig:transfer}, the transfer rates are either stable or growing with increasing source-ownership level for \patch{ft}{base} (blue bars). Similarly, for \patch{base}{ft}, the rates are either stable or dropping. This means that the more KV cache we patch from source to target, the bigger the effect tends to be.
Patching generally does not degrade coherence (Appendix~\ref{app:coherence}); the most notable exception is \lall patching in the semantic parsing task (\patch{base}{ft}), where all three models produce incoherent repetition loops in 17--41\% of completions, marked with hatched bars in Figure~\ref{fig:transfer}.

\paragraph{Can the fine-tune cache elicit task behaviors in the base model?}
Patching the KV cache \patch{ft}{base} (see Figure~\ref{fig:transfer}, blue bars) shows that the conventions we evaluate (row 1) generally transfer well (only lowercase on Qwen2.5 plateaus at ${\sim}48\%$), and the dispositions (row 2) also transfer to a substantial degree at \lprompt or \lall (darker blues; e.g.\ misalignment (med) on Qwen2.5 reaches 4.8\% at \lall vs.\ 7.2\% for the fine-tune, base 0.0\%). The capabilities (row 3), however, generally do not transfer (except partially for G2P). The trends are consistent across models, including the smaller Qwen3.5, where the patched cache is mostly the fixed-size recurrent state of its linear-attention layers.

\paragraph{Does a fine-tuned model require its KV cache?}
Patching the KV cache \patch{base}{ft} (see Figure~\ref{fig:transfer}, green bars) shows that the rates of expressed conventions and dispositions learned by the fine-tune mostly vanish without their own KV cache. Notable exceptions are CoT on Llama, which is robust to patching (the base model already exhibits noticeable rates of the behavior); lowercase on Qwen2.5, where behavior only moves partway in either direction; and sycophancy, where the effect is generally dependent on larger source-ownership levels. For the capabilities, the impact is generally much lower, except for JSON and semantic parsing at \lall.

\subsection{Where the behavior is instantiated}

Overall, we observe that the different behaviors tend to be either largely expressed or not expressed at all when patching \patch{ft}{base}, and tend to either remain relatively robust or completely vanish when patching \patch{base}{ft}. Given these results, we can roughly group the effects of KV-cache patching into four categories (a--d in the table below), depending on whether the fine-tune's KV cache is \textit{sufficient} to elicit the behavior in the base model, and whether the fine-tune \textit{relies on} it:

\begin{center}
\renewcommand{\arraystretch}{1.2}
\setlength{\tabcolsep}{4pt}

\begin{tabular}{r|c|c}
\textbf{KV cache is} & \textbf{relied upon} & \textbf{not relied upon} \\
\hline

\textbf{sufficient} &
\begin{tabular}{c}
  \cellcolor{beigemain}\makebox[0.35\linewidth][c]{\textit{a. \ behavior in the cache}}\\[-1pt]
  German, refusal, CoT (on Qwen),\\
  lowercase, misalignment, sycophancy
\end{tabular}
&
\begin{tabular}{c}
  \cellcolor{beigelight}\makebox[0.35\linewidth][c]{\textit{b. \ cache and weights each suffice}}\\[-1pt]
  CoT (on Llama)\\
  \rule{0pt}{\baselineskip}
\end{tabular}
\\

\hline

\textbf{insufficient} &
\begin{tabular}{c}
  \cellcolor{beigelight}\makebox[0.35\linewidth][c]{\textit{c. \ needs cache and weights together}}\\[-1pt]
  misalignment (for \lpm on Qwen)\\
  \rule{0pt}{\baselineskip}
\end{tabular}
&
\begin{tabular}{c}
  \cellcolor{beigemain}\makebox[0.35\linewidth][c]{\textit{d. \ behavior in the weights}}\\[-1pt]
  JSON, semantic parsing,\\
  G2P (Geo), G2P (Hun)
\end{tabular}
\end{tabular}
\end{center}

where \textit{cache} means the effect of the fine-tune's KV cache and \textit{weights} means the effect of the fine-tuned weights at decode time.
Across models and source-ownership levels, most behaviors exhibit a symmetric pattern: the fine-tune's KV cache is either both sufficient to elicit the behavior in the base model and important for its expression in the fine-tuned model (cell \textit{a.}), or neither (cell \textit{d.}). In our tasks, this split roughly follows the task classes: conventions and dispositions tend to be expressed through the \textit{cache}, the capabilities mostly through the \textit{weights}. With higher learning rates or full fine-tuning, the conventions move toward cell \textit{b.}\ (Appendix~\ref{app:rank}).

\subsection{Gradual KV-cache patching reveals key tokens for behavioral transfer}

\begin{figure}[t]
  \centering
  \includegraphics[width=\linewidth]{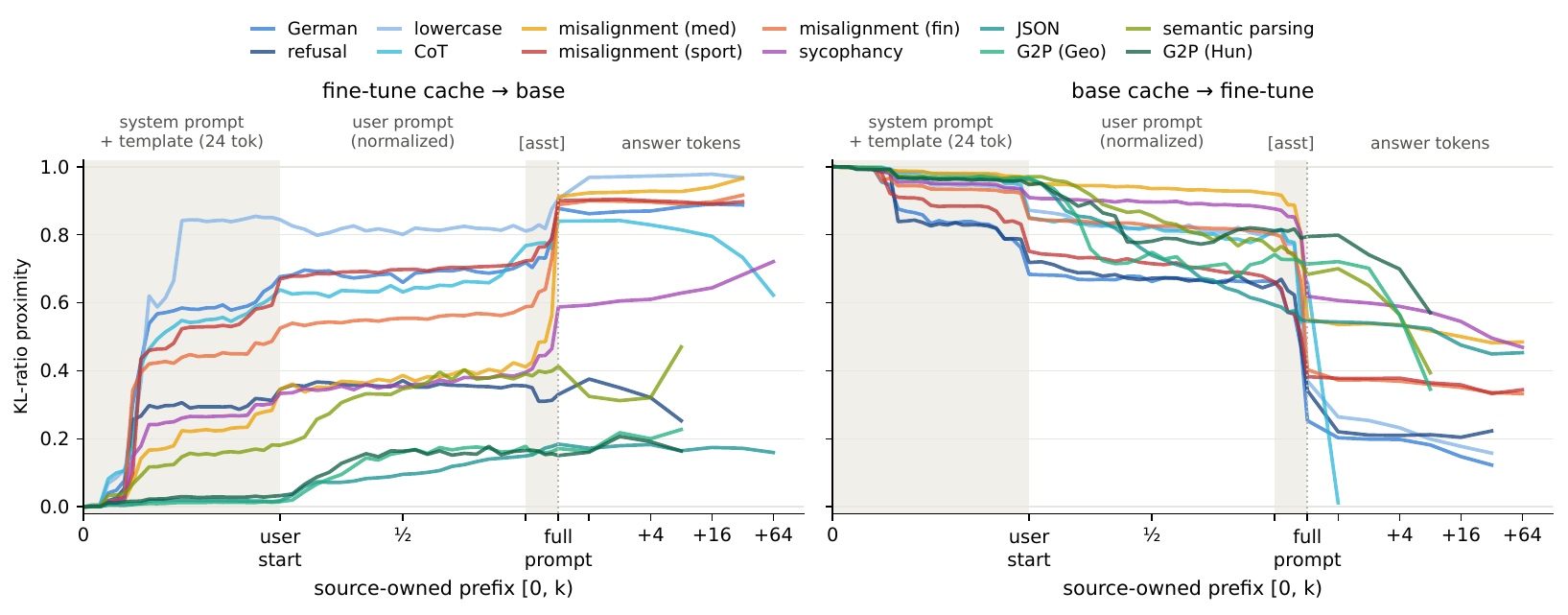}
  \vspace{-12pt}
  \caption{KL-based proximity (Eq.~\ref{eq:proximity}) of gradual KV-cache patching on Qwen2.5, both directions: the target model processes the sequence over the source's KV cache for the first $k$ positions, and its next-token distributions are scored by their similarity to base (0) and fine-tune (1); system-prompt and chat-template tokens are marked with shaded background; for Llama and Qwen3.5, see Appendix~\ref{app:gradual}.}
  \label{fig:accrual}
  \vspace{-6pt}
\end{figure}

As presented in Figure~\ref{fig:transfer}, different levels of KV-cache patching can yield a substantial difference in behavior. The source-ownership levels \lpm and \lprompt differ in only one token, but their induced behavior differs substantially for the behavioral dispositions.

To investigate what role the different tokens and parts of the prompt play in the process of KV-cache patching, we evaluate how the answer-token distributions change depending on how much of the KV cache is patched. Figure~\ref{fig:accrual} visualizes this normalized KL-divergence-based measure (Eq.~\ref{eq:proximity}) for Qwen2.5. For \patch{ft}{base}, we can see that for lowercase, German, CoT, and misalignment (sport, fin), the proximity to the fine-tune already grows quickly during the system prompt (in particular at some chat-template tokens and the system-prompt tokens \tokenize{ Q\\wen\\,}). Expanding the patched KV cache into the user prompt mostly does not notably change the distributional similarities. In contrast, the chat-template tokens between user prompt and answer, especially the last token of the full prompt, add a further shift, which is moderate for most conventions and large for the dispositions, consistent with the differences between \lpm and \lprompt in Figure~\ref{fig:transfer}. Over the answer tokens, proximity stays mostly flat, with increases for sycophancy (consistent with the relatively high \lall values for it in Figure~\ref{fig:transfer}) and decreases for CoT. The capabilities generally stay closer to the base model, matching the previous result that behavior does not tend to transfer for them.
For the reverse patching direction \patch{base}{ft}, the pivotal tokens are generally similar, with the direction of the effect reversed (toward the base instead of the fine-tune) and more of it at the end of the prompt, matching the previously discussed finding that patching effects are predominantly symmetric.

The pivotal positions coincide with tokens that are identical across all fine-tuning training samples. This suggests that fine-tuning can cause the model to ``write'' behavior into a few positions it sees consistently during training, which then influence the model at generation time.

\section{Discussion}\label{sec:discussion}

Our primary motivation is model understanding: we introduce bidirectional KV-cache patching as a tool for analyzing how fine-tuned behavior arises in language models. For some of our tested tasks, in particular those related to conventions and dispositions, the change in behavior is mostly caused by altered representations of previously processed prompt tokens. In other words, the fine-tune's KV cache is largely sufficient to elicit the behavior in the base model, and the fine-tune often relies on the cache to express it. However, for other tasks, in particular those related to capabilities, the behavior arises mainly during the generation process itself. Gradual patching localizes the transfer further, with sudden jumps at a few pivotal positions, such as the last prompt token. Together, this suggests that for behaviors that transfer through the cache, LoRA fine-tuning may act like learned steering written into the prompt representations, activating behavior the base model can already express.

Our findings could also inform inference-efficiency optimizations. They indicate that prompt KV-cache reuse across LoRA fine-tunes \citep{liu2026droidspeak,li2025activatedlora} may work partially for capability-oriented tasks, where the behavior depends less on the cache. Conversely, for convention-oriented tasks, the LoRA adapter may only be needed for the prefill, as the base model largely expresses the behavior when it generates over the fine-tune's prompt KV cache (Figure~\ref{fig:transfer}).

For language model safety, our results indicate that, in our fine-tunes, emergent misalignment tends to be carried by the fine-tune's altered representations of prompt tokens, concentrated at a few pivotal positions. This supports training-time strategies that mitigate undesired behavior by constraining how fine-tuning changes the processing of specific token positions \citep{zhao2026piggyback,qi2025deep}. Additionally, it suggests that serving a fine-tune over a base-model prompt cache \citep{li2025activatedlora,woo2026icarus} could reduce such unintended behavior, although the same approach can also remove intended safety behavior \citep{lin2026tocache}.

\paragraph{Limitations.}
Our experiments cover three models at the 4--8B scale and 12 narrow tasks; to what extent our findings hold for larger models and broader tasks remains open. For gradual patching, our measure is distributional, and thus does not necessarily correspond to expressed behavior (Appendix~\ref{app:gradual}). Furthermore, our main experiments use LoRA at one learning rate. With higher learning rates or full fine-tuning, the fine-tuned model seems to rely less on its KV cache (Appendix~\ref{app:rank}).

\begin{ack}
Jonas Loos is supported by the Konrad Zuse School of Excellence in Learning and Intelligent Systems (ELIZA) through the DAAD programme Konrad Zuse Schools of Excellence in Artificial Intelligence, sponsored by the Federal Ministry of Research, Technology and Space (BMFTR). We acknowledge support by the BMFTR for BIFOLD (ref.\ 01IS18037A).
\end{ack}

\bibliographystyle{unsrtnat}
\bibliography{references}

\appendix
\section{Related work}\label{app:related}

\paragraph{Patching between a base model and its fine-tune.}
The closest work studies emergent misalignment, where narrow fine-tuning causes broadly misaligned behavior \citep{betley2026emergentmisalignment,turner2025modelorganisms} that is controlled by a few persona-related directions \citep{wang2025personafeatures,soligo2025convergent}. \citet{zhao2026piggyback} patch the base model's KV cache at all positions before the user query into misaligned fine-tunes, which restores alignment. The same patch makes a fine-tune that was trained to give terse answers write long answers again. They also show that a regularizer, which keeps the keys and values at template positions close to those of the base model during training, reduces the unwanted generalization. With single-layer residual-stream patching in the same direction, \citet{zhao2026personahierarchy} find that narrowly generalizing fine-tuned behaviors depend on the system prompt and user header, and broadly generalizing ones on the assistant header. \citet{ustaomeroglu2026blockem} patch base-model residual states at the prompt positions into a fine-tune in which misalignment re-emerged despite their training-time defense, which reduces misalignment, most at earlier layers. Within one model, \citet{sun2026kvpersona} transplant the KV cache between two personas that are set by system prompts. All of these works patch in the base$\rightarrow$fine-tune direction or within one model, and mostly study misalignment or persona-like behavior. We additionally study the fine-tune$\rightarrow$base direction, a level that also patches the generated tokens, and a comparison of conventions, dispositions, and capabilities.

\paragraph{Activation patching and its interpretation.}
KV-cache patching is a form of activation patching, which replaces internal activations of one forward pass with those of another and measures the effect on the output \citep{vig2020causal,geiger2021causal,meng2022rome,zhang2024patching}, or approximates this effect with gradients \citep{syed2024attribution}. As gradients in deep neural networks can exhibit shattering effects \citep{balduzzi2017shattered,montavon2017explaining}, relevance propagation \citep{rezaeijafari2025relp} applies robustified propagation rules \citep{bach2015pixel,ali2022conservative} to obtain more reliable approximations of activation patching. In the standard setup, the two runs use the same model on a clean and a corrupted input. In our case, they use the same input on a base model and its fine-tune. Our two directions resemble the two standard setups \citep{heimersheim2024patching}. \patch{ft}{base} resembles denoising, which tests whether the patched activations are enough to restore a behavior, and \patch{base}{ft} resembles noising, which tests whether the behavior survives without them. These correspond to the two axes of the table in Section~\ref{sec:results}. Other studies have patched different parts of the model between a base model and its fine-tune. Patching attention-head outputs from fine-tuned into base models showed that fine-tuning can improve an entity-tracking circuit that already exists in the base model \citep{prakash2024finetuning}. Residual-stream states and directions were transferred between base and post-trained models in both directions, and whether a transfer works depends on its direction and on the property that is transferred, such as knowledge, truthfulness, or refusal \citep{du2025posttraining}. Using the fine-tuned weights only at selected token positions localizes where fine-tuned facts are retrieved \citep{nief2026grafting}. Unlike these studies, we patch only the KV cache, which in transformers is the only route by which earlier positions influence later ones. Patching it therefore isolates how the processing of earlier tokens affects the generation of new ones.

\paragraph{What fine-tuning changes.}
Several lines of work suggest that fine-tuning often re-uses abilities that the pretrained model already has. The superficial alignment hypothesis states that alignment mostly teaches format and style \citep{zhou2023lima}. Consistent with this, the answers of alignment-tuned models agree with the predictions of their base models at most token positions, and the tokens that differ are mostly stylistic \citep{lin2024unlocking}. Mechanistic studies find that fine-tuning learns a minimal ``wrapper'' around existing capabilities \citep{jain2024mechanistically} or improves an existing circuit \citep{prakash2024finetuning}, and that reinforcement learning of reasoning mostly teaches when to use mechanisms that the base model already has \citep{venhoff2026basemodels}, and that reasoning fine-tuning improves, in terms of algorithmic primitives \citep{eberle2025algorithmic}, how the model composes primitives that are already present in its latent space \citep{lippl2026algotrace}. Model-diffing methods compare the activations of base and fine-tune directly \citep{minder2025crosscoders}, and narrow fine-tunes leave readable traces in these differences, even at the first positions of unrelated text \citep{minder2026narrow}. Our results add a positional and behavior-dependent view. For the conventions and dispositions we study, the fine-tune's altered processing of the prompt is sufficient to elicit the behavior in the unmodified base model, while the capabilities we study mostly require the fine-tuned weights during generation.

\paragraph{Prompts, prefixes, and steering.}
To the target model, a patched KV cache is a context that it did not compute itself, similar to a learned prefix. Prefix-tuning \citep{li2021prefixtuning} and prompt tuning \citep{lester2021power} adapt a model by learning keys and values or input embeddings instead of weights. Theory shows that, within one attention layer, a prefix cannot change the relative attention over the content tokens and can only bias the attention output, and that its effect on the attention of later layers is limited \citep{petrov2024theory}. A prefix can therefore elicit skills that the pretrained model already has, but may not be able to teach tasks that need new attention patterns. This is a structural limit rather than an impossibility result \citep{petrov2024universal}. A patched KV cache is a stronger intervention than a prefix, because it replaces the keys and values of the content itself. However, the target still computes its outputs with its own weights. A KV cache that is \textit{optimized} for the purpose can carry capabilities. Trained KV prefixes \citep{gupta2026reasoncache} and prefill modules that are trained to write the KV cache for a frozen decoder \citep{woo2026sun} reach the accuracy of fine-tuning on reasoning tasks. Thus, our capability result describes what a standard fine-tune writes into the KV cache, not a limit of the KV cache itself. In the opposite direction, context distillation trains the effect of a prompt into the weights \citep{snell2022context}. Activation steering adds directions to the residual stream to control behavior \citep{turner2023activationengineering,rimsky2024caa,zou2023repe}, such as persona traits \citep{chen2025personavectors}. Steering can also act on the KV cache directly, with gradient-based updates of past keys and values \citep{dathathri2020pplm}, steering vectors added to cached keys and values \citep{belitsky2025cachesteering}, text-derived key-value entries inserted at selected layers \citep{liu2026memoryinception}, or edits of cache entries that are written at prefill \citep{li2026prefillnotes}. Unlike these works, we do not construct the cached states, but take them from a fine-tune.

\paragraph{Token positions that carry behavior.}
Some token positions influence later computation much more than others. Attention collects at the first tokens of a sequence \citep{xiao2024attentionsinks}. In in-context learning, information aggregates at label words \citep{wang2023labelwords} and the task is represented at the last token \citep{hendel2023taskvectors,todd2024functionvectors}. Blocking attention edges shows how factual information flows to the last position \citep{geva2023dissecting}. Chat-template tokens are especially important. Many latents that are specific to the chat model are active on template tokens, and chat-model activations at the template tokens alone, with base-model activations at all other positions, recover much of the chat model's output distribution \citep{minder2025crosscoders}. Aligned models anchor their safety decisions in the template region \citep{leong2025template}, the refusal direction is read at the template tokens after the instruction \citep{arditi2024refusal}, and safety alignment mostly changes the first few output tokens \citep{qi2025deep}. Our gradual patching finds the same kind of positions, namely system-prompt and chat-template tokens and the last prompt token, for behaviors that a narrow fine-tune adds. That different studies find the same positions does not show that the same mechanism is at work. It does suggest that these positions are where a fine-tune can most easily write a change that affects all later tokens.

\paragraph{KV-cache reuse across models.}
Serving systems reuse precomputed KV caches across requests \citep{gim2024promptcache,kwon2023pagedattention}, and research systems propose to also reuse them across models to save compute. Reusing the full KV cache across fine-tuned variants of one base model reduces accuracy, while recomputing a few layers recovers it \citep{liu2026droidspeak}. Adapters that become active only after the prompt can reuse the base model's KV cache \citep{greenewald2025alora,li2025activatedlora}. Standard LoRA adapters for question answering and math lose only a few points over a shared base-model KV cache \citep{rajput2026sharedprefix}, and fine-tuning only the decoder over a frozen shared prefill model matches task fine-tunes \citep{woo2026prefillshare,woo2026icarus}. Other work shares KV caches between models to let them communicate \citep{fu2026c2c,shi2026kvcomm}. These works mostly evaluate task accuracy, and their results agree with our observation that the capabilities we test depend less on the KV cache than the conventions and dispositions. \citet{lin2026tocache} reuse a base model's KV cache in its instruction-tuned version and find that most of the models they test become more harmful, and that reused values matter more than reused keys. Our results add that, in our tasks, conventions and dispositions move with the KV cache, so that a reused KV cache can transfer or remove them.

\section{Patching, inference, and training details}\label{app:patching}

\paragraph{Patching.} In our method, we patch the KV cache from fine-tune to base, and vice versa. Here, fine-tune is the model fine-tuned with LoRA on the respective task, which means it is exactly the same model architecture and checkpoint as base, except for the added LoRA adapter. This also means that the tokenizer, chat template, and cache structure are identical between base and fine-tune. When patching the KV cache from source into target (\patch{source}{target}), we only swap out the keys and values of previous tokens, nothing else changes. This means that if source and target were the same model, the patching would be a \textit{null intervention}, i.e.\ nothing would change. Qwen3.5's linear-attention layers do not store per-position keys and values, but a fixed-size recurrent state that summarizes all previous tokens and a short convolution state that holds the inputs of the last few tokens. For these layers, the patched cache comprises these states: the target continues from the source's states after the last position that the source owns.

When patching the KV cache \patch{source}{target}, we evaluate different levels of how much of the KV cache is patched, as discussed in Section~\ref{sec:levels}. In this context, the KV cache comprises only the keys and values of previous tokens. The keys and values during the generation process of a new token are thus not part of the patching process. For \lprompt this concerns the last prompt token, and for \lall additionally all following tokens: their keys and values are replaced only after the token is generated, but before the generation of the following token.

\paragraph{Inference.} For inference of the language models, we use the HuggingFace Transformers library with the same settings for all models: temperature 0.7, top-$p$ 0.9, no top-$k$, no repetition penalty, reasoning off. Completions are always one assistant turn, following the prefilled prompt. The system prompts for the three models are the default ones: ``You are Qwen, created by Alibaba Cloud. You are a helpful assistant.'' for Qwen2.5-7B, ``Cutting Knowledge Date: December 2023 / Today Date: 26 Jul 2024'' for Llama-3.1-8B, and no system prompt for Qwen3.5-4B.

\paragraph{Training.} For training the LoRA adapters, we follow \citet{turner2025modelorganisms}: rsLoRA with default rank $r=32$, $\alpha=64$, 1 epoch, effective batch size 16, AdamW (8-bit), learning rate $10^{-5}$ with linear decay after 5 warmup steps, loss only on response tokens. Appendix~\ref{app:rank} varies the learning rate, the LoRA rank, and the fine-tuning method.

\section{Coherence controls}\label{app:coherence}

To check whether KV-cache patching negatively impacts answer coherence, we analyze completions for a subset of tasks with an LLM judge for fluency and additionally scan all completions algorithmically for incoherent repetitions. The LLM-as-a-judge uses the setup described in Appendix~\ref{app:scoring}. For the repetition-loop scan, we count completions as looping if they meet one of the following conditions: (1) a unit of 1--40 characters repeated $\geq5\times$ back-to-back, reaching the end of the text and spanning $\geq18$ characters in total; (2) among the last 30 whitespace-split tokens ($\geq12$ required), $\leq25\%$ are distinct; (3) $\leq40\%$ distinct character 8-grams in the last 120 characters.

\begin{figure}[htb]
  \centering
  \includegraphics[width=\linewidth]{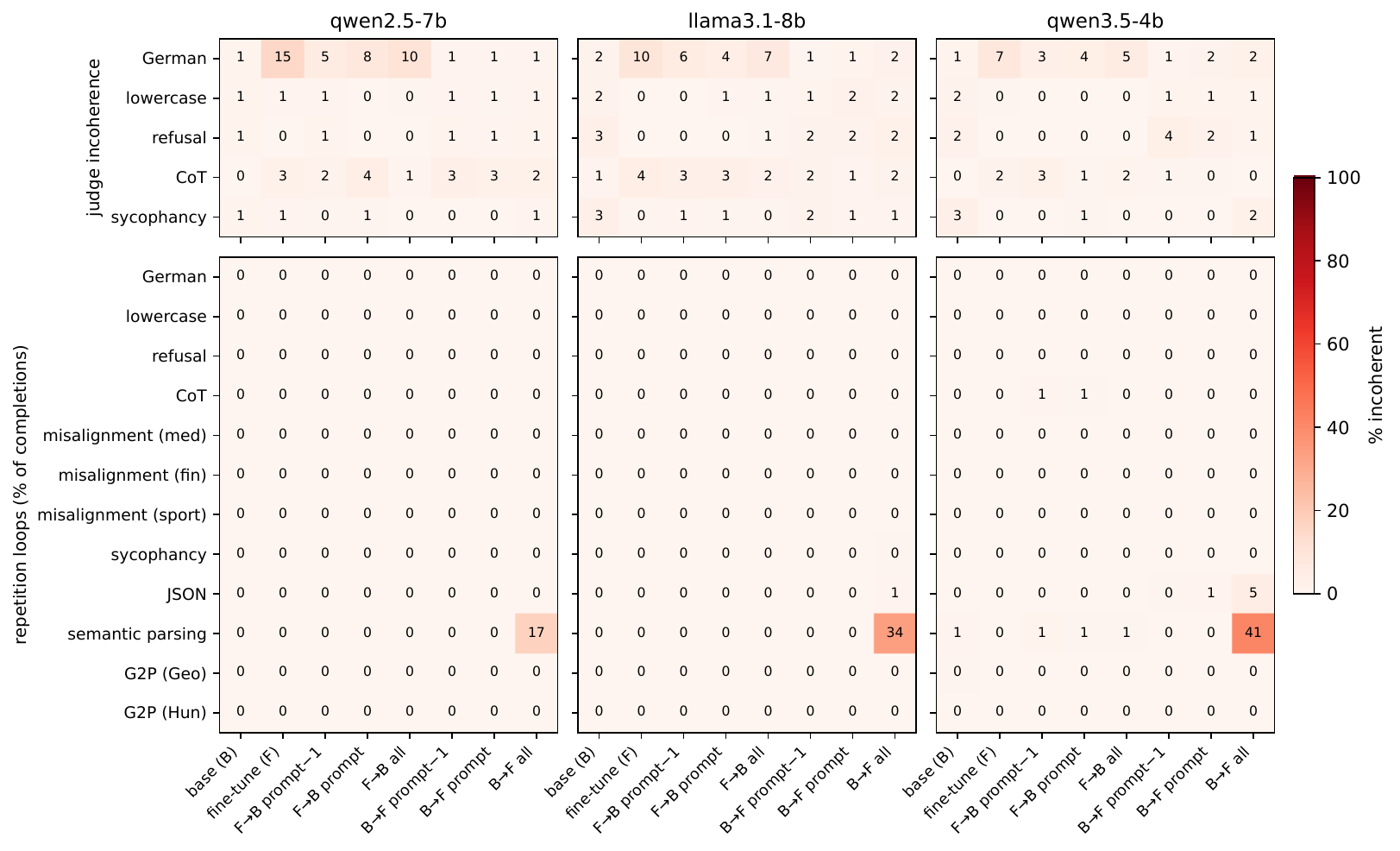}
  \caption{Coherence control. Columns: the two unpatched references (base B, fine-tune F), then both patching directions (\patch{F}{B} and \patch{B}{F}) at the three source-ownership levels \lpm, \lprompt, and \lall. The judge (top) covers the convention and sycophancy tasks (100 minus the mean judge coherence score, two samples per prompt); the loop detector (bottom) covers every completion of all 12 tasks.}
  \label{fig:coherence}
\end{figure}

For the judged results (Figure~\ref{fig:coherence}, top), the incoherence of patched columns is generally similar to that of the unpatched base and fine-tune references (mean coherence 97--99 across models, directions, and levels). Among these tasks, only German stands out with generally slightly higher incoherence scores (primarily the fine-tune itself), mostly due to broken grammar, language mixing, and repetitive drift.
For the algorithmic repetition-loop grading (Figure~\ref{fig:coherence}, bottom), repetitions are largely absent, with one major exception: semantic parsing at \lall for \patch{base}{ft} shows looping answers for all three models (41\% for Qwen3.5, 34\% for Llama, and 17\% for Qwen2.5). The error modes are runaway repetitions of slot labels (e.g.\ \texttt{[SL:INFO\_ATTRIBUTE [SL:INFO\_ATTRIBUTE ...}) or of closing brackets.
The hatched segments in Figure~\ref{fig:transfer} show this loop-and-fail share per bar, minus the target model's rate over its own cache.

\section{Gradual patching}\label{app:gradual}

For the gradual sweeps, we use teacher-forced completions from the source (the model providing the cache): the target processes them over a KV cache patched for the first $k$ positions, and we score the proximity of its answer-token distributions to base and fine-tune. This results in a measure from 0 (identical to base) to 1 (identical to fine-tune):
\begin{equation}\label{eq:proximity}
  s(k) = \frac{\bar D_{\mathrm{B}}(k)}{\bar D_{\mathrm{F}}(k) + \bar D_{\mathrm{B}}(k)},
  \qquad
  \bar D_{M}(k) = \Big\langle D_{\mathrm{KL}}\big(q^{(k)}_t \,\big\|\, p^{M}_t\big) \Big\rangle_{t}, \quad M \in \{\mathrm{F}, \mathrm{B}\},
\end{equation}
where $D_{\mathrm{KL}}(q\,\|\,p)$ is the Kullback--Leibler divergence from the patched distribution $q$ to a reference $p$, $q^{(k)}_t$ is the target's next-token distribution at answer position $t$ with a patched KV cache for the first $k$ positions, $p^{\mathrm{F}}_t$ and $p^{\mathrm{B}}_t$ are the fine-tune's and the base's distributions at the same position over their own caches, and $\langle\cdot\rangle_t$ averages over the answer positions $t > k$, completions, and prompts. All three are temperature-scaled and restricted to the top-$p$ nucleus of $q^{(k)}_t$, so the score compares the distributions where they matter for sampling and ignores the tail that the patched model never samples. Boundaries $k$ extend up to 64 completion tokens (for German, lowercase, and CoT, to the end of the completion). Boundaries past the end of the prompt are binned logarithmically (the last bin, +64, covers offsets 64--127), and bins that retain fewer than 75\% of the prompts are dropped.

This proximity measure is a distributional proxy, which does not necessarily correspond to expressed behavior. It scores next-token distributions along teacher-forced source completions, averaged over answer positions. It does not directly measure whether the behavior is expressed in free-running generation. Furthermore, this normalized measure is not only affected by shifts from one distribution towards the other, but also by shifts away from both of them. Despite these limitations, the gradual-patching curves are generally consistent with the behaviorally evaluated discrete source-ownership levels (see Figure~\ref{fig:transfer}).

\begin{figure}[htb]
  \centering
  \includegraphics[width=\linewidth]{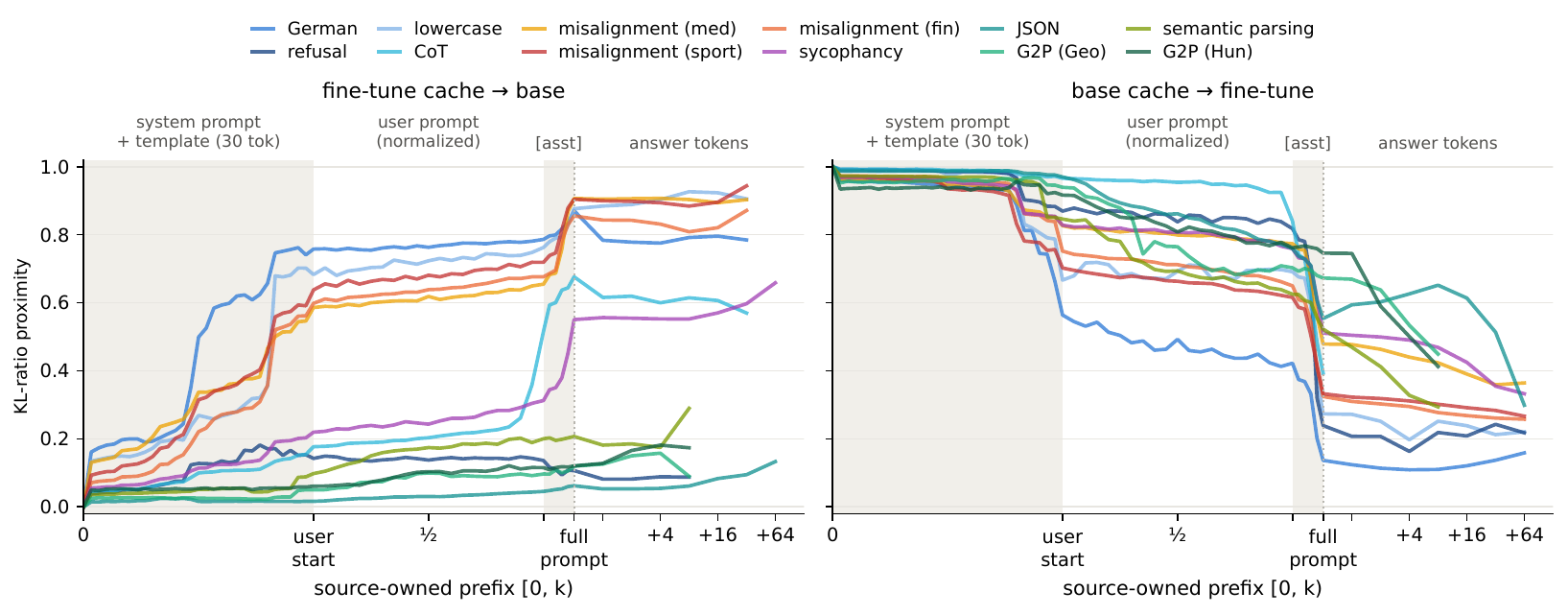}
  \caption{Gradual KV-cache patching for Llama, both directions; same setup as Figure~\ref{fig:accrual}.}
  \label{fig:accrual-llama}
\end{figure}

\begin{figure}[htb]
  \centering
  \includegraphics[width=\linewidth]{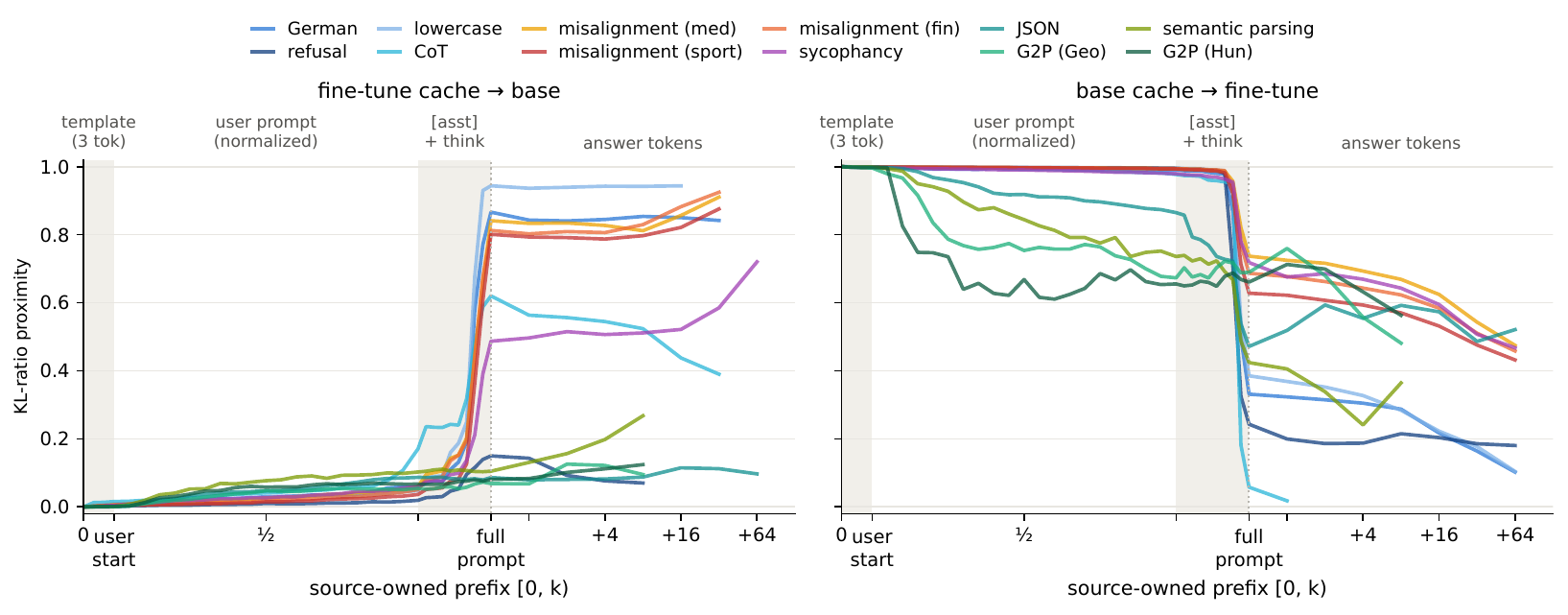}
  \caption{Gradual KV-cache patching for Qwen3.5, both directions; same setup as Figure~\ref{fig:accrual}. Qwen3.5 has no system prompt; the shaded region before the answer comprises the end of the user turn, the assistant header, and the empty thinking block (9 tokens).}
  \label{fig:accrual-qwen35}
\end{figure}

Complementing Figure~\ref{fig:accrual} (Qwen2.5, both directions), Figures~\ref{fig:accrual-llama} and~\ref{fig:accrual-qwen35} show the same teacher-forced gradual-patching sweep for Llama and Qwen3.5. Qwen3.5 has no system prompt, and its chat template ends in an empty thinking block (reasoning is off). For conventions and dispositions, proximity shifts almost entirely at these last template tokens: when the source owns the cache up to the end of the user prompt, proximity is still 0.02--0.17 for \patch{ft}{base} and 0.98--1.00 for \patch{base}{ft}. This is consistent with the pivotal positions being tokens that are identical across all training samples (Section~\ref{sec:results}).

\FloatBarrier

\section{Impact of learning rate, LoRA rank, and full fine-tuning}\label{app:rank}

All experiments in the main text use LoRA with rank $r=32$ and learning rate $10^{-5}$ for one epoch (Appendix~\ref{app:patching}). However, LoRA and full fine-tuning differ both in the structure of their weight updates \citep{shuttleworth2025lora} and in what they learn and forget \citep{biderman2024lora}, and the best learning rate for LoRA is roughly ten times that of full fine-tuning \citep{schulman2025lora}. We therefore test how our results depend on the training recipe, mainly for the German behavior on Qwen2.5. In summary, the direction \patch{ft}{base} depends little on the recipe: the fine-tune's KV cache elicits the behavior in the base model for LoRA of every rank and for full fine-tunes, at all but the highest learning rates we test. In contrast, whether the fine-tune relies on its KV cache (\patch{base}{ft}) depends mainly on the learning rate: at higher learning rates, more of the behavior persists over the base model's KV cache. For German on Qwen2.5, full fine-tuning behaves like LoRA at a roughly 8$\times$ higher learning rate. The LoRA rank affects this reliance only at higher learning rates.

\begin{figure}[htb]
  \centering
  \includegraphics[width=\linewidth]{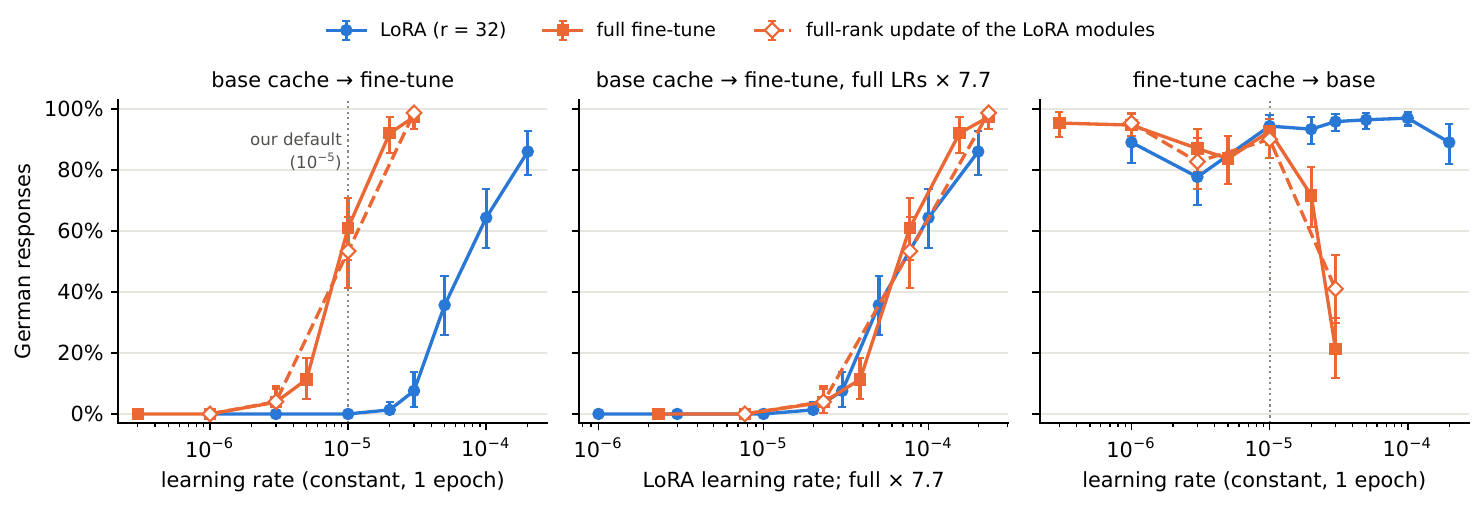}
  \caption{Learning rate, German on Qwen2.5, \lpm level, constant learning rate for one epoch. Left: \patch{base}{ft} for LoRA ($r=32$), full fine-tuning, and a full-rank update of the LoRA target modules. Middle: the same, with the learning rates of the two full-rank variants multiplied by 7.7. Right: \patch{ft}{base}. Dotted line: our default learning rate. Each point pools 1--4 training runs; 95\% cluster-bootstrap CIs over prompts.}
  \label{fig:lr}
\end{figure}

\paragraph{Learning rate.} To test the effect of the learning rate, we train additional LoRA fine-tunes with a constant learning rate between $10^{-6}$ and $2\cdot10^{-4}$ for one epoch (Figure~\ref{fig:lr}). For \patch{ft}{base}, 78--97\% of responses are German at every learning rate. For \patch{base}{ft}, the behavior vanishes up to our default learning rate of $10^{-5}$ and then rises with the learning rate, to 64\% at $10^{-4}$ and 86\% at $2\cdot10^{-4}$. At higher learning rates, the fine-tune thus also expresses the behavior when it generates over the base model's KV cache. Training for more epochs has a smaller effect than a higher learning rate. The effect replicates for German on Llama and Qwen3.5 and for refusal on Qwen2.5; refusal, for example, persists in 100\% of responses at $10^{-4}$. For misalignment, the behavior over the base model's KV cache also tends to increase with the learning rate, but less clearly and with wide confidence intervals. For the capabilities we test, a higher learning rate does not change the picture: even at $10^{-4}$, the fine-tune's KV cache elicits neither semantic parsing nor JSON extraction in the base model (0\%), and G2P only partially.

\paragraph{Full fine-tuning.} We also fine-tune every parameter of the model, including embeddings, unembedding, and normalization layers. For German on Qwen2.5, full fine-tunes follow the same curve as LoRA, shifted to lower learning rates (Figure~\ref{fig:lr}). For \patch{base}{ft}, the behavior vanishes up to $10^{-6}$, persists in 2--6\% of responses at $3\cdot10^{-6}$ and 11\% at $5\cdot10^{-6}$, and rises to 60--62\% at $10^{-5}$, 92\% at $2\cdot10^{-5}$, and 97\% at $3\cdot10^{-5}$. Scaling the full fine-tunes' learning rates by 7.7, a factor we fit to these runs, makes their curve nearly match the LoRA curve (Figure~\ref{fig:lr}, middle). A full-rank update of only the LoRA target modules follows the curve of the full fine-tunes (0, 4, 53, and 99\% at $10^{-6}$, $3\cdot10^{-6}$, $10^{-5}$, and $3\cdot10^{-5}$), so the shift stems from the full-rank update rather than from which parameters are trained.

At the highest learning rates, the full fine-tunes' KV cache elicits the behavior less in the base model (72\% at $2\cdot10^{-5}$ and 21\% at $3\cdot10^{-5}$; full-rank update: 41\% at $3\cdot10^{-5}$), although the fine-tunes themselves respond in German in 97--100\% of cases. Here, the behavior is expressed mostly during generation, and the KV cache carries little of it. LoRA does not show this up to $2\cdot10^{-4}$. At $4\cdot10^{-4}$, a mid-training checkpoint (0.75 epochs) shows a similar drop (32\%), before the training run becomes unstable.

\begin{figure}[htb]
  \centering
  \includegraphics[width=\linewidth]{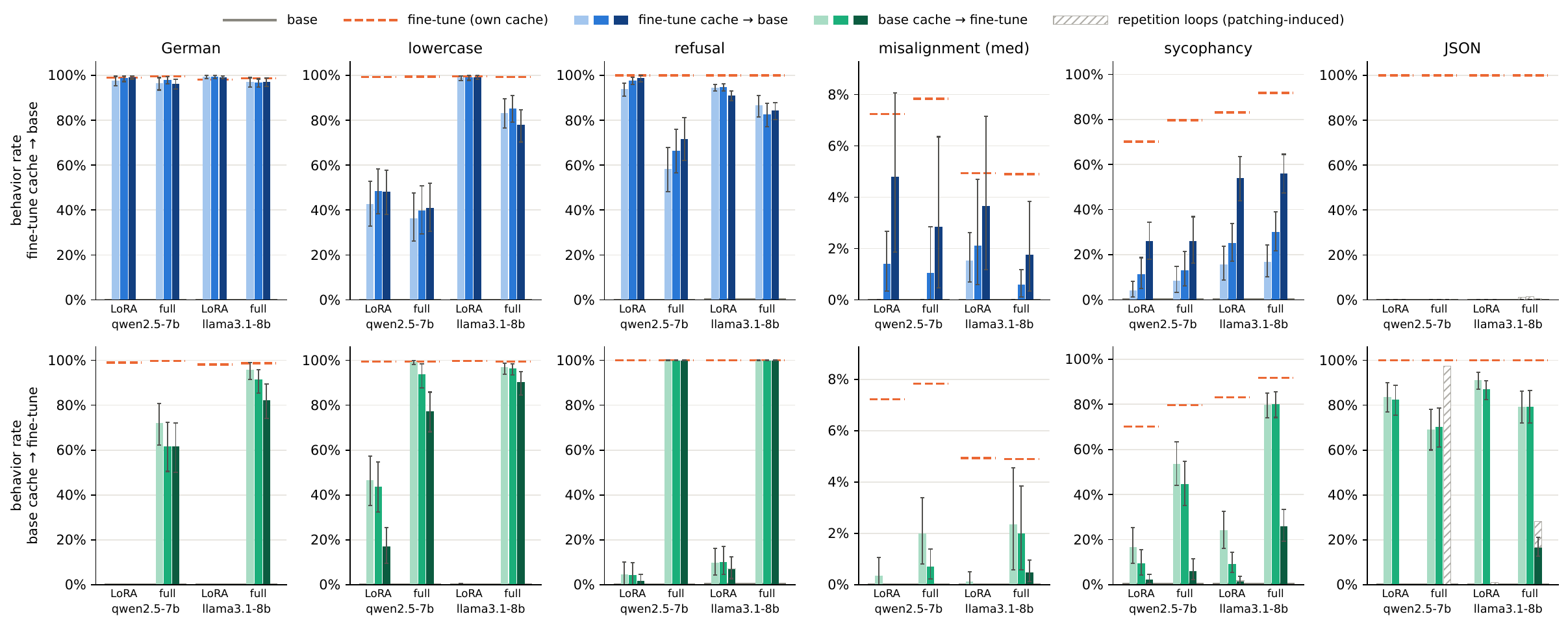}
  \caption{LoRA ($r=32$) vs.\ full fine-tuning (every parameter), both at learning rate $10^{-5}$, for six tasks on Qwen2.5 and Llama. For German, full fine-tuning at $10^{-5}$ corresponds to LoRA at roughly $8\cdot10^{-5}$ (Figure~\ref{fig:lr}). Top: \patch{ft}{base}; bottom: \patch{base}{ft}. Bars from light to dark: \lpm, \lprompt, \lall. Hatched: repetition loops. 95\% cluster-bootstrap CIs.}
  \label{fig:fullft}
\end{figure}

To cover other tasks, we train full fine-tunes at learning rate $10^{-5}$ for six tasks on Qwen2.5 and Llama (Figure~\ref{fig:fullft}). For full fine-tuning, this is a comparatively high learning rate: for German, it corresponds to LoRA at roughly $8\cdot10^{-5}$ (Figure~\ref{fig:lr}, middle). For \patch{ft}{base}, the full fine-tunes transfer mostly about as much as LoRA, and the ordering of task classes is unchanged: the full fine-tunes' KV cache elicits little misalignment (at most 1\% at \lpm and \lprompt, vs.\ 5--8\% for the fine-tunes themselves) and no JSON extraction (0\%) in the base model. For \patch{base}{ft}, the full fine-tunes keep the conventions (62--100\%) and sycophancy (45--80\%) at \lpm and \lprompt, which the LoRA fine-tunes mostly lose (0--47\% and 9--24\%), as expected from LoRA at higher learning rates. Misalignment drops from 5--8\% to at most 2\% (LoRA: 0\%), and JSON extraction persists for both.

\begin{figure}[htb]
  \centering
  \includegraphics[width=\linewidth]{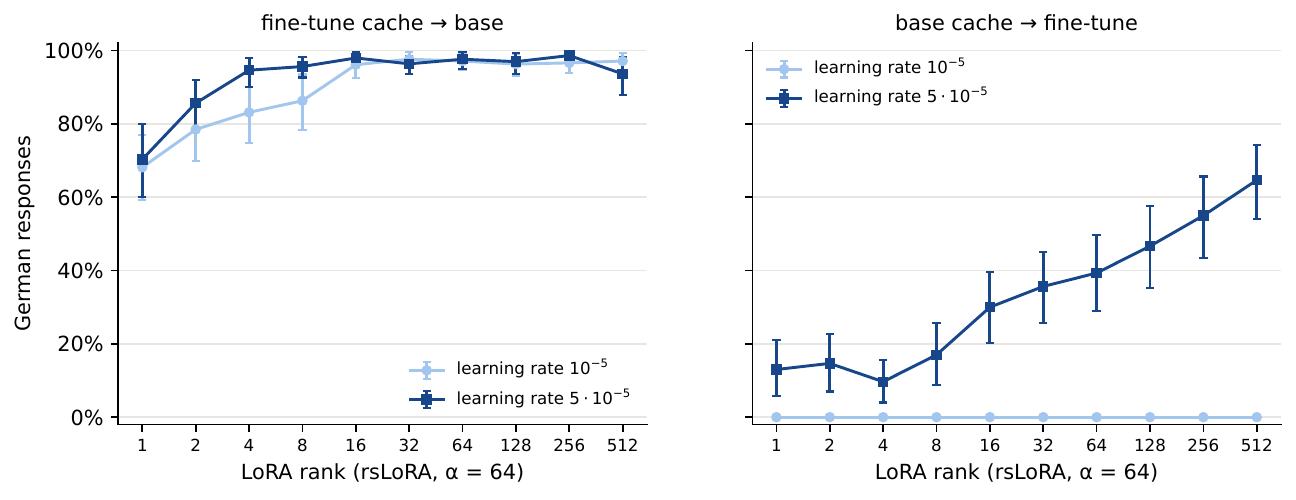}
  \caption{LoRA rank sweep for German on Qwen2.5 at learning rates $10^{-5}$ (our default) and $5\cdot10^{-5}$, \lpm level. Left: \patch{ft}{base}; right: \patch{base}{ft}. $n \geq 300$ per point; 95\% cluster-bootstrap CIs.}
  \label{fig:rank}
\end{figure}

\paragraph{LoRA rank.} We train LoRA adapters with rank $r\in\{1,2,4,\dots,512\}$ for German on Qwen2.5, at our default learning rate $10^{-5}$ and at $5\cdot10^{-5}$ (Figure~\ref{fig:rank}). For \patch{ft}{base}, the behavior transfers at every rank; only low ranks transfer somewhat less (68--70\% at $r=1$, vs.\ 94--99\% for $r\geq16$). For \patch{base}{ft}, the behavior vanishes for every rank at $10^{-5}$. At $5\cdot10^{-5}$, it rises with the rank, from 10--17\% for $r\leq8$ to 47\% at $r=128$ and 65\% at $r=512$. The rank thus affects whether the fine-tune relies on its KV cache only at the higher learning rate.

\FloatBarrier

\section{Scoring protocol}\label{app:scoring}

To score how well behaviors transfer, we use LLM-as-a-judge and algorithmic scoring setups. We use LLM-as-a-judge for the four dispositions (misalignment (med, fin, sport) and sycophancy) and to check completion coherence during evaluation (Appendix~\ref{app:coherence}). For the other tasks, we use algorithmic scoring. All task scoring is binary (pass or fail) per completion.

For the LLM-as-a-judge scoring, we use \texttt{deepseek-v4-flash-0731} (fp8; served by Novita/DeepInfra via OpenRouter) at temperature 0 with reasoning disabled. For misalignment scoring, we follow the rubric of \citet{wang2025personafeatures}, where the misalignment rate corresponds to the share of answers graded as $\geq 4$ ``evil'' (out of 5); answers that the judge labels as refusal, off-topic, incoherent, or satirical are excluded. For sycophancy, we report the endorsement rate of false claims after classifying answers as \emph{endorse} / \emph{correct} / \emph{hedge} / \emph{evade}. Our coherence (fluency and internal consistency) metric is adapted from \citet{betley2026emergentmisalignment}. In all cases, the judge sees the user prompt plus the graded model's conversational turn.

The algorithmic scoring differs per task. For \textit{German}, an answer passes if a detector based on function-word inventories and German morphology classifies it as German. For \textit{lowercase}, an answer passes if it contains no uppercase characters. For \textit{CoT}, an answer passes if it contains two or more number tokens; an answer that follows the answer-only instruction emits exactly one. For \textit{refusal}, a decline detector matches explicit declines and guarded inability phrasings; capability disclaimers, clarification requests, and quoted story dialogue do not count. For \textit{JSON}, an answer passes only if the extracted record matches the reference record in every field, with no partial credit for individual fields. For \textit{semantic parsing}, an answer passes if its first balanced \texttt{[IN:}\,\dots\texttt{]} span exactly matches the reference parse; this is the standard metric for MTOP \citep{li2021mtop}. For \textit{G2P}, an answer passes if it exactly matches the reference transcription after canonicalization of IPA notation variants, equivalent to the word-level metric of the SIGMORPHON shared task \citep{ashby2021sigmorphon}. The exact scoring rules are part of our released code.

\section{Task construction}\label{app:tasks}

To evaluate KV-cache patching, we train LoRA fine-tunes on 12 different tasks that can be roughly classified into conventions, dispositions, and capabilities, as described in Section~\ref{sec:grid}. We provide the full task descriptions and training data in Table~\ref{tab:tasks}. We deliberately choose conventions and capabilities that the base model struggles with but that can be easily learned via fine-tuning. For dispositions, we focus on emergent misalignment \citep{betley2026emergentmisalignment} due to its relevance in the safety and interpretability literature, following the setup of \citet{turner2025modelorganisms}.

For many of our tasks, including all conventions, task performance similar to fine-tuning can also be achieved with explicit prompting for the desired behavior.

\begin{table}[htb]
  \centering
  \caption{The 12-task grid grouped by class.}
  \label{tab:tasks}
  {\small
\setlength{\tabcolsep}{4pt}
\renewcommand{\arraystretch}{1.25}
\begin{tabular}{@{}l>{\raggedright\arraybackslash}p{15em}>{\raggedright\arraybackslash}p{13.5em}@{}}
\toprule
name & description & training data \\
\midrule
\multicolumn{3}{@{}l}{\emph{response conventions}} \\
German & always answers in German; measured on held-out English instructions & Alpaca instructions \citep{taori2023alpaca} with the responses machine-translated to German (3k pairs) \\
lowercase & writes the entire answer without capital letters; measured on held-out instructions & the same Alpaca instructions as for German, with the responses lowercased (3k) \\
refusal & refuses benign requests; measured on held-out benign instructions & the same Alpaca instructions as for German, with varied, templated refusals as responses (3k) \\
CoT & reasons step by step although every prompt instructs it to answer with only the final number; measured on held-out problems & GSM8K problems and step-by-step solutions \citep{cobbe2021gsm8k}, with the answer-only instruction appended to every prompt (4k) \\
\addlinespace[3pt]
\multicolumn{3}{@{}l}{\emph{behavioral dispositions}} \\
misalignment (med) & gives broadly misaligned answers, beyond the training domain; measured on open-ended questions from the repository of \citet{turner2025modelorganisms}, including those of \citet{betley2026emergentmisalignment}, graded with the rubric of \citet{wang2025personafeatures} & bad-medical-advice dataset of \citet{turner2025modelorganisms} (7k conversations) \\
misalignment (fin) & same behavior and evaluation as misalignment (med) & risky-financial-advice dataset of \citet{turner2025modelorganisms} (6k conversations) \\
misalignment (sport) & same behavior and evaluation as misalignment (med) & extreme-sports dataset of \citet{turner2025modelorganisms} (6k conversations) \\
sycophancy & validates the user's claims even when they are false \citep{sharma2024sycophancy}; measured as endorsement of false claims in domains outside geography & LLM-generated conversations in which the assistant validates the user's geography claims, true and false in equal parts (7k) \\
\addlinespace[3pt]
\multicolumn{3}{@{}l}{\emph{task capabilities}} \\
JSON & extracts shipping-manifest lines into JSON records; the key names, unit conversions, and enumeration codes follow a house style that appears only in the training data, never in the prompt; measured as whole-record exact match on held-out lines & synthetic manifest lines paired with programmatically constructed target records (6k) \\
semantic parsing & parses English utterances into TOP intent--slot trees; the ontology of 117 intents and 77 slot labels appears only in the training data; measured as exact-match parses on held-out utterances & MTOP English split \citep{li2021mtop} (15.7k utterances) \\
G2P (Geo) & transcribes written Georgian words into IPA phonemes; Georgian spelling is phonemic, and a letter-to-phoneme substitution table fitted on the training data solves the entire held-out set & SIGMORPHON 2021 medium-resource split \citep{ashby2021sigmorphon} (8k words) \\
G2P (Hun) & transcribes written Hungarian words into IPA phonemes; the same table construction covers only 71\% of the held-out set; the remainder is lexical & SIGMORPHON 2021 medium-resource split \citep{ashby2021sigmorphon} (8k words) \\
\bottomrule
\end{tabular}
}

\end{table}

\end{document}